# Time series model selection with a meta-learning approach; evidence from a pool of forecasting algorithms

Sasan Barak[1,2,*], Mahdi Nasiri[3], Mehrdad Rostamzadeh[4]

1. Department of Management Science, Lancaster University Management School, Lancaster University, UK
2. Faculty of Economics, Technical University of Ostrava, Ostrava, Czech Republic
3. Department of Physics, Sharif University of Technology, Tehran, Iran. Email: m.nasiri@physics.sharif.edu
4. Department of Industrial Engineering, Sharif University of Technology, Tehran, Iran. Email: m.rostamzadeh@ie.sharif.edu

*Corresponding Author: Sasan Barak,
Address: Faculty of Economics, Sokolská třída 33, Ostrava, Czech Republic.,
Tel.: +420720495247; fax: +420597 322 008.
E-mail addresses: s.barak@lancaster.ac.uk, sasan.barak@vsb.cz

## Abstract:

One of the challenging questions in time series forecasting is how to find the best algorithm. In recent years, a recommender system scheme has been developed for time series analysis using a meta-learning approach. This system selects the best forecasting method with consideration of the time series characteristics. In this paper, we propose a novel approach to focusing on some of the unanswered questions resulting from the use of meta-learning in time series forecasting. Therefore, three main gaps in previous works are addressed including, analyzing various subsets of top forecasters as inputs for meta-learners; evaluating the effect of forecasting error measures; and assessing the role of the dimensionality of the feature space on the forecasting errors of meta-learners. All of these objectives are achieved with the help of a diverse state-of-the-art pool of forecasters and meta-learners. For this purpose, first, a pool of forecasting algorithms is implemented on the NN5 competition dataset and ranked based on the two error measures. Then, six machine-learning classifiers known as meta-learners, are trained on the extracted features of the time series in order to assign the most suitable forecasting method for the various subsets of the pool of forecasters. Furthermore, two-dimensionality reduction methods are implemented in order to investigate the role of feature space dimension on the performance of meta-learners. In general, it was found that meta-learners were able to defeat all of the individual benchmark forecasters; this performance was improved even after applying the feature selection method.

## 1. Introduction

Time series forecasting has been a vital part of business analytics for many years; from forecasting the financial markets, to forecasting European beverage consumption (Heravi, Osborn, & Birchenhall, 2004). In recent decades, numerous methods have been developed by different groups in the hope of achieving the best method for forecasting any time series with any horizon. However, it has been understood from the "No free lunch" theorem (Wolpert & Macready, 1997), that there is no single method able to achieve the best performance for every time series it faces.

In order to select a model for forecasting a time series, two methods are accustomed in the forecasting community: aggregate selection and individual selection. In aggregate selection, one model is selected to forecast all of the time series, while in the individual selection for each of the time series, the best performer model is selected (Fildes, 1989). To select the best performer for each time series, two major approaches have been developed: information criteria and empirical accuracy. In the first scheme, penalised maximum likelihood methods such as Akaike information criteria (AIC) and its extensions, are used to choose the best model among the candidate models for a specified time series. In the second approach, with the help of cross-validation and error measures, the best performer model is selected as the proper model for the time series (R. J. Hyndman & Koehler, 2006; Qi & Zhang, 2001).

One of the most significant downsides of these model selection methodologies is the computational cost, as it is usual to test many forecasters on all of the data. If the size of the dataset at hand exceeds a certain threshold, the computational burden will prove to be unjustifiable. These problems have led the forecasting community to search for alternatives.

One promising replacement to the previous model selection approaches is creating a recommendation system using machine-learning algorithms for the time series forecasting, which are capable of suggesting the most powerful forecasting method for a given time series (Arinze, 1994; Meade, 2000). Prudêncio and Ludermir (2004), suggested a scheme in which firstly, different forecasters are implemented on a set of time series and a ranking of the performances are created for the data. Next, a set of characteristics called the meta-features are extracted from the time series and with the help of the ranking, a set of classifiers known as the meta-learners are trained on this data. Finally, the trained meta-learners are asked to suggest the best forecasters for the future time series. These recommendation systems can greatly decrease the computational costs of large sets of time series, by discarding the need to implement all forecasters on a time series to find the most suitable one.

Many attempts have been made to improve the results of using meta-learners as recommender systems for time series analysis. Shah (1997), used discriminant analysis to select the best forecasting method for time series analysis among a group of three forecasting methods. Wang, Smith-Miles, and Hyndman (2009), used supervised and unsupervised learning methods and generated recommendation rules to select the most suitable forecasting method for the time series based on their meta-features. Lemke and Gabrys (2010), defined a set of features for the time series and analysed the results of applying different meta-learning approaches to the time series of the NN3 and NN5 datasets. Widodo and Budi (2013), proposed a model to select the proper forecasting method in which they applied principal component analysis to reduce the dimension of their data. Kück, Crone, and Freitag (2016), studied the use of different feature sets for the neural network meta-learner. They also used error-based features and statistical tests as meta-features. Talagala, Hyndman, and Athanasopoulos (2018), proposed a general framework to select the best forecasting algorithm for time series using the random forest classifier.

Despite the numerous studies conducted in this field, there are still many questions unanswered regarding the parameters affecting the meta-learners in time series analysis. The first major issue in this field is the problem of the meta-feature pool. Each research has used its own set of features based on the performances and the results gained from a particular dataset, and there is no clear justification or pattern for defining the suitable set of features.

One other key factor which was not addressed in previous works is the role of various error measures. The issue of considering different error measures for analysing the performance of forecasters was always contentious in the forecasting community (Davydenko & Fildes, 2016). The important aspect of this problem for our study, is the role of different error measures in the performances of meta-learning algorithms since each ranking from the meta-learning model selection, is defined based on a unique error measure.

Another important unanswered question is that regarding the pool of forecasters used as the target attributes for the meta-learning algorithms. Two significant aspects of this pool – the number and quality of the forecasters, are not addressed independently in previous studies. It is accustomed in the forecasting community that each study defines their pool in a way that supports their intended result; however, with this approach of arbitrary forecaster pools, the effects of a different set of forecasters on meta-learners cannot be targeted. It is of great interest to address the properties of forecaster pools since the quality of input data can vary drastically by adding forecasters with non-consistent performances, and this quality can affect the performances and errors of different meta-learning algorithms. If a pattern is discoverable, it can help the forecasting community to design standard meta-learning criteria for every dataset.

This work aims to focus on the gaps of previous studies (as mentioned above), by proposing a novel approach to studying the inputs of meta-learning algorithms. This objective will be reached by introducing a diverse state-of-the-art pool of forecasters and meta-learners, and also analysing three key concepts:

- Evaluation of using various subsets of forecasters as inputs for meta-learners
- Assessing the effectiveness of feature selection on the meta-learning approach
- Consideration of the robustness of meta-learning in different error measures

Since direct applications and performances of meta-learning based methods are important for industry-level practitioners, at the end of this work we will analyse the performances of our meta-learners on a common pool of forecasters in the business known as 'basic forecasters'.

The rest of this paper is structured as follows: in section 2, a background of all of the methods and algorithms used in this research is provided. In section 3, the methodology of the procedures and novelties implemented in this paper are presented. In section 4, the detailed results of our studies are provided followed by a discussion in section 5. Finally, section 6 concludes the paper.

## 2. Background

In this section, we will provide a technical description of the algorithms and methods used in this study.

## 2.1 Forecasting methods

The forecasting methods used in this paper are all statistical methods; from the *M3 competition* (Makridakis & Hibon, 2000) winner *theta* method, to the popular *exponential smoothing*.

*2.1.1 Naïve & Seasonal Naive Methods*

In the Naïve method, the most recent observation is used for all forecasts. The h-step ahead forecast is simply set to be the value of the last observation. In time series notation:

$$\hat{Y}_{t+h|t} = Y_t \qquad (1)$$

$Y_t$ denotes a time series and $\hat{Y}_{t+h|t}$ is the h-step ahead forecast (R. J. Hyndman & Athanasopoulos, 2018). In seasonal Naïve, the forecasts are set to be equal to the last observed value of the same season of the data. In this approach the h-step ahead forecast is written as:

$$\hat{Y}_{t+h|t} = Y_{t+h-s(k+1)} \qquad (2)$$

$$K = \frac{\lfloor h-1 \rfloor}{s} + 1 \qquad (3)$$

where *s* denotes the seasonal period. The implementation of these methods were achieved by the *naive()* and *snaive()* functions in the R package *forecast* (R. J. Hyndman & Khandakar, 2007).

*2.1.2 Exponential Smoothing*

This relatively old model was first proposed by Brown (1959), and is one of the building blocks of many modern algorithms for time series forecasting. Within these methods, four components of a time series (level, trend, seasonality, and damp) are used in a multiplicative or additive way to model the whole time series (Winters, 1960). The state space representation of the parameters of exponential smoothing models was developed by R. Hyndman, Koehler, Ord, and Snyder (2008), which is generally referred to as ETS. In this paper 'ZZZ' model refers to the best ETS model selected automatically by AIC as implemented from the R package *smooth* (Svetunkov, 2018).

2.1.3 Simple moving average (SMA)

In this method, the h-step ahead forecast is set equal to the average of the historical observations. That is:

$$\hat{Y}_{t+h|t} = \frac{Y_1 + Y_2 + \cdots + Y_t}{t} \qquad (4)$$

where $\hat{Y}_{t+h|t}$ is an estimation of time series $Y$ at time $t+h$ (R. J. Hyndman & Athanasopoulos, 2018).

*2.1.4 Multiple temporal aggregation algorithm*

The multiple temporal aggregation algorithm (MAPA) was suggested by N. Kourentzes, Petropoulos, and Trapero (2014), with the motivation of precisely targeting time series' components with the help of multiple temporal structures. The ETS models were used in the first MAPA versions as the forecasters needed these for each temporal component, and the resulting components (trend, level, seasonality) were aggregated to create the final model. More details on different temporal aggregation methods and their performances can be found in (Barrow & Kourentzes, 2016; N. Kourentzes et al., 2014).

In this paper, the implementation of this algorithm was achieved via the R package *MAPA* (N. Kourentzes, F. Petropoulos, 2018). Here the frequency of time series sets is equal to 7, since the data was recorded on a daily basis, and the mean was used as the combination operator. In each aggregation level, the type of ETS used is 'ZZZ'.

*2.1.5 Temporal Hierarchies (Thief)*

This method uses contiguous hierarchical temporal aggregations to model the time series. At each level, predictions can be made with any forecasting method (unlike earlier versions of MAPA where only exponential smoothing models were used), and then the values are aggregated by the proposed structure. More details on the temporal hierarchies forecasting performance can be found in (Athanasopoulos, Hyndman, Kourentzes, & Petropoulos, 2017). In this study, the implementation of the temporal hierarchies on the data was achieved via the *thief* package in the R language (Hyndman RJ, 2018). For the *thief()* function, the seasonal period was set to 7, the combination method of temporal hierarchies was set to structural scaling ('struc'), and the model used for forecasting each aggregation level was set to 'theta'.

*2.1.6 TBATS*

This method was first suggested by De Livera, Hyndman, and Snyder (2011), to address some of the problems such as complex seasonal patterns and calendar effects which classical exponential smoothing models could not apprehend. The model is named after the key components comprising of its algorithms: trigonometric seasonal modes, Box-Cox transformation, ARMA errors and trend and seasonal components. We have used the *forecast* package in the R language (R. J. Hyndman & Khandakar, 2007) to deploy the TBATS method in this study. The ARMA errors were considered in the *tbats()* function, and then the best fit was selected by the Akaike information criterion.

*2.1.7 Theta Method*

Theta method is a univariate forecasting approach applied in forecasting a non-seasonal time series. In this method, a new time series is obtained by solving a second-order difference equation by decomposing the original time series into lines called 'Theta lines'. Each of these lines is extrapolated with a forecasting algorithm, and the forecasts are combined in order to obtain the forecast for the main time series (Assimakopoulos & Nikolopoulos, 2000). The implementation of this algorithm was achieved via the R package *forecTheta* (Jose Augusto Fiorucci, Louzada, Yiqi, & Fiorucci, 2016).

*2.1.8 Dynamic optimized theta method (DOTM)*

Dynamic optimized theta method is a modified version of the theta method that produces forecasts which unlike the standard theta method are not necessarily linear. This method is introduced by Jose A Fiorucci, Pellegrini, Louzada, Petropoulos, and Koehler (2016). Here again, the R package *forecTheta* was used for the implementation.

## 2.2 Error Measures

### 2.2.1 sMAPE

Mean absolute percentage error (MAPE) is a percentage-based error that can be calculated as follows:

$$MAPE = mean\left(100\left|\frac{Y_t - \widehat{Y_t}}{Y_t}\right|\right) \quad (5)$$

Although MAPE is a popular forecasting error measure, it has a percentage error-based nature. As a result, it is prone to infinite values if the series contains zero values. The second disadvantage is the asymmetry of putting weights on forecasts; more details on this can be found in Armstrong and Collopy (1992).

With the MAPE deficiencies in mind, it is preferable to use a modified version of MAPE. This version is called symmetric mean absolute percentage error (sMAPE) and can be calculated as follows:

$$sMAPE = mean\left(200\frac{|Y_t - \widehat{Y_t}|}{|Y_t + \widehat{Y_t}|}\right) \quad (6)$$

In the equations above, $Y_t$ is the actual value at time t; whilst $\widehat{Y_t}$ is the forecast value at the given period (R. J. Hyndman & Koehler, 2006).

### 2.2.2 MASE

Mean absolute scaled error which was first introduced by R. J. Hyndman and Koehler (2006) is a scaled measure created for non-seasonal and seasonal time series. For a non-seasonal time series, MASE can be defined as

$$MASE = mean\left(\left|\frac{e_i}{\frac{1}{T-1}\sum_{i=2}^{T}|Y_t - Y_{t-1}|}\right|\right) \quad (7)$$

where $e_i$ in the nominator is the forecasting error at the given period and the denominator calculates the mean of the in-sample forecast error ($MAE$) for the naïve forecasting algorithm.

For a seasonal time series, MASE can be calculated as:

$$seasonal\ MASE = mean\left(\left|\frac{e_i}{\frac{1}{T-d}\sum_{t=m+1}^{T}|Y_t - Y_{t-m}|}\right|\right) \quad (8)$$

Similar to the non-seasonal case, $e_i$ in the nominator is the forecast error at the given period, $Y_i$ represents the actual value of the series at the given period, and $d$ is the seasonal period. In the denominator, the in-sample forecast error ($MAE$) for the seasonal naïve algorithm is calculated (R. J. Hyndman & Athanasopoulos, 2018). Here in this paper, the seasonal MASE is used as the second error measure.

## 2.3 Meta-features for time series

One of the earliest ideas in using features for time series classifying was proposed by Nanopoulos, Alcock, and Manolopoulos (2001). Later, this idea was extended by Wang, Smith, and Hyndman (2006), to cluster the time series based on meta-features. In recent years, each research has developed its own set of features for time series analysis (Fulcher & Jones, 2014; Kang, Belušić, & Smith-Miles, 2014; Wang et al., 2009).

The structure of using the features for forecasting the time series in this paper will follow that of Kang, Hyndman, and Smith-Miles (2017).

## 2.4 Meta-learners

Meta-learning was first suggested by Rice (1976), to improve the results of the learning algorithms by finding the best scheme for the problem. Rice named this process 'algorithm selection problem' (ASP). Later, a formal description for meta learning was proposed by Smith-Miles (2009), for a given problem with a given feature set; and the goal is to find selection mapping for the algorithm space – to select an algorithm that has maximum performance. Finding the selection mapping between the feature space and problem space for obtaining the best performance, is done by meta-learners which are mainly classification methods.

In this paper the following classifiers were used as meta-learners:

### *2.4.1 Artificial neural networks (ANN)*

Neural networks are non-parametric estimators that can be used for accurately estimating non-linear functions (Mehrotra, Mohan, & Ranka, 1997). The inputs of a neural network are vectors of variables corresponding to an observation. These vectors are weighted and combined with linear filters and become the inputs of hidden layers where non-linear computation called activation function is performed on inputs, and calculates the output of the network (Gershenfeld, 1999). The whole process of learning is achieved by adjusting the weight parameters, and the network is updated each time it has been fed new data. After the parameters are updated, the desired outcome will form the classification of our data.

### *2.4.2 Decision Tree*

The decision tree algorithm is a non-parametric and non-linear machine learning algorithm used for regression and classification problems. This algorithm takes advantage of a hierarchical structure for recursively segmenting training data in order to build a suitable model. The most common strategy in constructing a decision tree is a top-down method which recursively partitions the data into subgroups until a termination criterion has been met. Determining termination criteria is a crucial task since it can prevent growing branches that do not affect the tree quality (Murthy, 1998).

### *2.4.3 Bagged Tree (Treebag)*

Bagging, which is a short term for 'bootstrap aggregation', is an ensemble algorithm that uses the bootstrap method to generate samples of the original data. For each sample data, a decision tree is constructed and the results are combined. For classification purposes, this algorithm averages the predicted values over a collection of bootstrap samples, and the class of a new observation is the most selected class among the number of trees constructed on bootstrap samples (Breiman, 1996; Kuhn & Johnson, 2013).

### *2.4.4 Random Forest (RF)*

Random forest is a type of ensemble classifier, made by a combination of decision trees to overcome any vulnerabilities of the single tree. For constructing an RF model, random samples are made with the help of the bootstrap method and then for each sample data, an unpruned decision tree is constructed. The difference between this method and the bagged tree is that in constructing each tree in RF, a subset of features is selected randomly to decrease the similarity of the trees (Archer & Kimes, 2008). Finally, the outcome forest of the trees is combined, and the average of the predictions are considered to be the result (Wang et al., 2018).

*2.4.5 Extreme gradient boosting (XGboost)*

XGboost is a non-linear machine learning algorithm used for functions such as classification, regression, and ranking (Chen, He, & Benesty, 2015). XGboost is an improved implementation of the gradient-boosted trees algorithm, in which a loss function is optimized by determining a set of parameters. More results about this algorithm can be found in Carmona, Climent, and Momparler (2018)

*2.4.6 Support Vector Machine (SVM)*

This algorithm converts the input data to a new higher dimensional space, called the feature space, using kernel functions. In the feature space, the equation for a hyperplane in this space is considered. Next, the cost function is computed, and with the help of Lagrange multipliers, the equation for the classifying hyperplane is written. More details can be found in Vapnik (2013).

Implementation of all classifiers was achieved by the *caret* package in the R language (Kuhn & Johnson, 2013).

## 2.5 Dimension Reduction methods

High dimensionality has always been a bottleneck for data analysis, and it has been previously shown that the reduction of this high value can significantly affect the performance of machine-learning methods (Aggarwal, 2001). In this paper, we used an unsupervised method, principal component analysis, and a feature selection method for the dimensionality reduction task. More details on the performances of different methods can be found in Van Der Maaten, Postma, and Van den Herik (2009).

*2.5.1 Principal component analysis*

Principal component analysis is one of the most well-known dimensionality reduction techniques invented by Pearson (1901). It projects high dimensional data into a lower dimensional space by extracting the main components of the data known as principal components (PCs). Principal components are uncorrelated and capture the possible variance of the main data (Zhong & Enke, 2017). The first $n$ principal components that represent a reasonable proportion of the variance, can be selected as new features that are descriptive for the main features (Kuhn & Johnson, 2013; Malhi & Gao, 2004).

*2.5.2 Feature selection*

One of the main differences of the feature selection method against the other dimensionality reduction algorithms is its ability to extract and save the pure form of the relevant data. Without any transformations like the ones involved in the PCA, feature selection can specify the redundant data, and this can directly target the future data collection processes (Khalid, Khalil, & Nasreen, 2014). In this paper, these methods were used to detect the most important set of features for the inputs of meta-learners.

## 3. Methodology
The goal of this paper is to address the following three main questions:

- What effects do different error measures (sMAPE, MASE) have on the meta-learners?
- What happens when different sets of forecasting methods are used as the outcome attribute (the 'best' label) for meta-learning? Is choosing the most powerful method always the perfect option for meta-learning?
- What is the role of the dimensionality of the feature space on the forecasting errors of meta-learners? Does feature selection improve or deteriorate the final results?

The whole work, as presented in Figure 1, can be divided into three phases. First, a set of time series are fed into different forecasting methods, and the ranking of the performances are listed based on the different forecasting error measures. With the help of this ranking, the forecaster with the lowest error measure is labelled as the 'best' forecaster for each of the time series.

In the second phase, the input data needed for the meta-learning algorithms are created. This task was achieved by combining the 'best' labels gained from the first phase with the meta-features extracted from the time series; the list of which is presented in Table 1. We will call this resulting new data-set as 'preliminary input'. The 'best' label will act as the classification label for the Meta learners.

After preparing the 'preliminary input', we could then focus on two of the main goals of this research: the effects of different forecasting methods on meta-learners, and the role of dimensionality. For this purpose, two new modifications were made on the 'preliminary input'. The first modification was achieved by considering three different subsets of top forecasters for determining the best label, this also meant the effects of different forecasters on the performances of meta-learners could be studied in more detail. The second modification was achieved by adjusting the dimensionality of the feature space by applying feature engineering methods such as feature selection and principle component analysis. These two new editions will result in a total of nine new versions of the 'preliminary input'. From now on these nine new versions will be known as the 'pool of inputs', as depicted with a dashed box in Figure 1.

In the third phase, the data from the pool of inputs is separated into training/test data. The recommender system for our algorithm will be the meta-leaners trained on the training data. Next, the trained models are requested to recommend the best method for each time series of the test data and the resulting errors of recommended methods are computed for the test data.

The pool of forecasting models considers four different sets of inputs to demonstrate the efficiency of the meta-learning in a different style of the preliminary models. It means that we first considered the top 4 and the top 6 forecasting models based on the error measure as the inputs pool. Later, to evaluate the effect of a lazy forecasting model on the meta-learning model selection, we also used Snaive besides the previous top 6 models, to make the third pool of forecasters.

Finally, to be able to validate and assess the performances of our recommender system on a general class of forecasters, the forth pool of models are trained on a set of forecasters relating to business-level applications. A family of exponential smoothing methods are used widely in business applications (R. Hyndman et al., 2008). In this paper, a separate alternative pool of forecasters is defined by considering four different members of the exponential smoothing family: ANN, ANA, AAN, and AAA. Hereafter, this new pool of forecasters will be known as 'basic' forecasters. Besides the popularity of these models in business applications, another motivation for defining this alternative pool of forecasters is the absence of any form of the model selection process (such as the AIC method in ZZZ) in these forecasting algorithms. This pure form of basic forecasters will help us to assess our model's performance on a more general scope of forecasters.

The pseudo code for our proposed framework is presented in Algorithm 1.

Table 1. List of features extracted from the time series

| Number | Features | Description |
|---|---|---|
| 1 | frequency | Frequency |
| 2 | nperiods | The number of seasonal periods in the series |
| 3 | seasonal_period1 | The length of the first seasonal period |
| 4 | seasonal_period2 | The length of the second seasonal period |
| 5 | trend | Trend |
| 6 | spike | Spike |
| 7 | linearity | Linearity |
| 8 | curvature | Curvature |
| 9 | e_acf1 | First ACF value of the remainder series |
| 10 | e_acf10 | The sum of the first 10 squared ACF values of the remaining series in a STL decomposition of the series. |
| 11 | seasonal_strength1 | The strength of the first seasonal component |
| 12 | seasonal_strength2 | The strength of the second seasonal component |
| 13 | peak1 | The location of the peak of the first seasonal component in STL decomposition |
| 14 | peak2 | The location of the peak of the second seasonal component in STL decomposition |
| 15 | trough1 | The location of the trough of the first seasonal component in STL decomposition |

| Number | Features | Description |
|---|---|---|
| 16 | trough2 | The location of the trough of the second seasonal component in STL decomposition |
| 17 | entropy | Spectral entropy |
| 18 | x_acf1 | The sum of the squared of first autocorrelation coefficients of the series |
| 19 | x_acf10 | The sum of the squared first ten autocorrelation coefficients of the series |
| 20 | diff1_acf1 | First ACF value of the differenced series |
| 21 | diff1_acf10 | Sum of squares of first ten autocorrelation coefficients of the first-differenced series |
| 22 | diff2_acf1 | First ACF value of twice differenced series |
| 23 | diff2_acf10 | Sum of squares of the first ten ACF values of the original series |
| 24 | seas_acf1 | The autocorrelation coefficient at the first seasonal lag |

## Algorithms 1. Forecasting based on meta learning

**First Phase**
**Given:**
  Collection of 111-time series: $[Y_1, Y_2, ..., Y_{111}]$

  Collection of 11+4 *basic* forecasting algorithms: $[A_1, A_2, ..., A_{15}]$
  The set of 2 forecasting error measures (MASE, sMAPE): $[M_1, M_2]$
  forecast horizon(H)
**Process**:
*For $j = 1$ to $15$*:
  *For $i = 1$ to $111$*:
    Handling the missing values for time series $[Y_i]$
    evaluating 3 rolling origins for time series $[Y_i]$ with $H = 56$
    forecast the time series $[Y_i]$ for h step ahead with algorithm $[A_j]$
      *For $k = [1, 2]$*:
        compute the forecast error measure $[M_k]$
      *end*
  *end*
  compute mean forecast error
*end*
Rank the forecasting algorithms among each group of main and "basic" forecasters based on the error measures
**Output:**
Mean forecast error of each forecasting algorithm
Ranking of main 11 forecasting algorithms based on error measures

**Second Phase**
**Given:**
Function to calculate the time series features
The set of different rankings (Top 4, 6, 6 + Snaive , "basic"): $[R_1, R_2, R_3, R_4]$
Feature engineering methods (Feature selection, PCA): $[S_1, S_2]$
**Process:**
*For $R = 1$ to $4$*:
  *For $K = [1, 2]$*:
    *For $i = 1$ to $111$*:
      Extract the features for time series $[Y_i]$
    *end*
    Add the "best" label for error measure $[M_k]$
  *end*
*end*

$For\ l = [1, 2]$:
    Apply the feature engineering methods $[S_l]$ only on the main dataset
$end$
**Output:**
The set of new inputs for meta-learners: $[V_1^{main}, V_2^{main}, ..., V_{111}^{main}]$ and $[, V_2^{basic}, ..., V_{111}^{basic}]$

**Third Phase**
**Given:**
The set of classification methods: $[c_1, c_2, ..., c_6]$
The set of new inputs divided as test/train data: $D_{train}, D_{test}$ for main and "basic" dataset
**Process:**
Identify the class labels
$For\ m = 1\ to\ 6$:
    $For\ t = 1\ to\ 5$:
        Create 10-fold cross-validated training data $[D_{train}^t]$
        Train the classifier $[c_m]$ with the new training set $[D_{train}^t]$
    $end$
    Predict the best forecasting method for the test set $[D_{test}]$
    Compute the forecasting errors for the recommended models
$end$
**Output:**
Meta learning forecast errors

## 4. Results

### 4.1 Dataset

Studies of this paper were conducted on the NN5 time series competition dataset, which consists of 111 empirical time series of daily cash withdrawals of ATMs in the UK (Andrawis, Atiya, & El-Shishiny, 2011). The objective of the NN5 competition was to obtain the lowest forecast error for 56 days ahead. Each time series contained 791 observations and included different patterns such as multiple seasonality, trends, structural breaks, zeros and missing values (Hung, Hung, & Lin, 2014). For data preprocessing, the missing values of each time series were replaced by the mean of the time series. The Box-Cox transformation was not applied to this dataset because of negligible effects ($\sim 0.001$ in forecasting accuracies).

### 4.2 Results of implementing individual forecasting errors

The first phase of the experiment was to forecast the 111-time series on the NN5 competition dataset. For this purpose, 111 time series with three rolling origins and forecasting horizons of 56 days ahead were considered, and with the help of the point forecasts, two error measures of sMAPE and MASE were computed. The final errors were achieved by the aggregation of the mean of all of the time series for each model and with the help of these errors, a ranking was created; the results of which can be found in Table 2. As it can be seen in Table 2, the MAPA method for the SMAPE and the MASE, outperformed the other forecasting algorithms. The results of the individual methods were used as a benchmark for our proposed meta-learning approach. However, since the implemented forecasting methods were not from the same class of family, using AIC and BIC as typical feature selection benchmarks were not considered, and so we only focused on the cross-validated error of the individual methods.

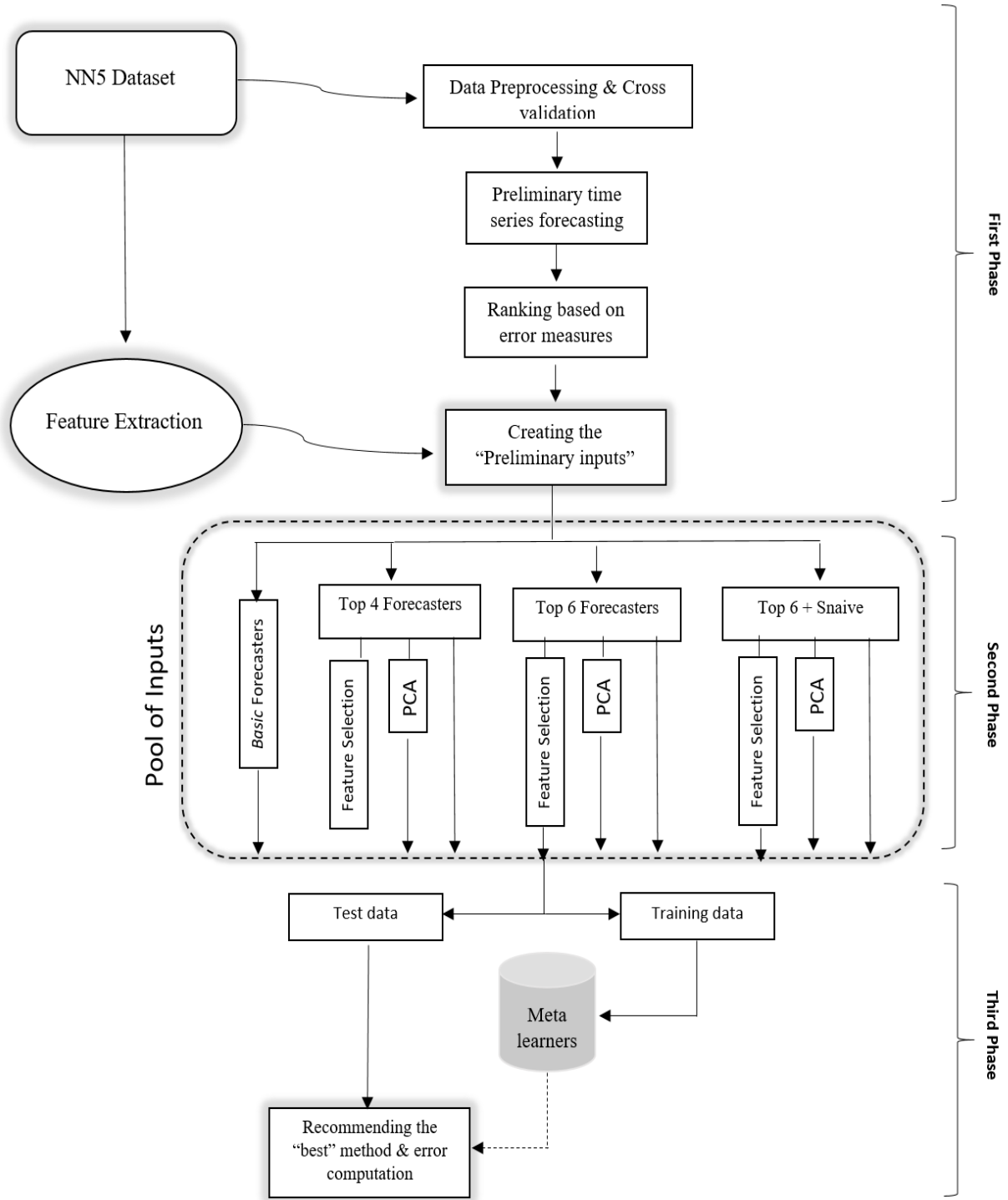

Figure 1. The proposed methodology of the paper

Table 2. Performances of different forecasting methods

| Model Name | sMAPE | MASE |
|---|---|---|
| **MAPA** | **0.213471** | **0.87706** |
| Temporal Hierarchy (Thief) | 0.214868 | 0.8815 |
| ZZZ | 0.21634 | 0.88521 |
| TBATS | 0.219058 | 0.888432 |
| Simple Theta | 0.219342 | 0.894925 |
| DOTM | 0.219553 | 0.896171 |
| Neural Net | 0.242186 | 0.995387 |
| ARIMA | 0.247171 | 0.890934 |
| Snaive | 0.25907 | 1 |
| SMA | 0.364972 | 1.652826 |
| Naive | 0.511456 | 2.289465 |

## 4.3 Meta features for time series

The second phase of the experiment was to create the pool of inputs for the meta learning algorithms. First, three main versions of the 'preliminary input' were formed by combining the set of features extracted from the time series with the different subsets of ranking labels (top 4, top 6, and top 6 with the seasonal naïve forecasting methods) gathered from the previous phase. The labels used here indicate the best method for each of the 111-time series, based on the desired error measure, using the rank of the method achieved in the previous phase. For the purpose of this research, we used a set of 24 features, which are presented in Table 1. More details about the feature sets and their computation can be found in (R. J. Hyndman, Wang, & Laptev, 2015; Kang et al., 2017). Extraction of the features from the NN5 dataset was achieved using the R language package *tsfeatures* (R. J. Hyndman, Wang, Kang, & Talagala, 2018). A complete covariance matrix of the feature set is provided in Figure 2.

## 4.4 Implementing feature selection

Next, the role of dimensionality on the input data was addressed. For dimension reduction, feature selection methods were applied to the input data; various feature selection methods such as *Info Gain, Gain Ratio, Chi-squared, Symmetric Uncertainty,* and *OneR*. Finally, the *OneR* (One Rule) algorithm was chosen for this work, because of its consistent performance on the various subsets of the input data (4,6, and 6 with Snaive). The *OneR* algorithm was implemented via the *Rapid miner* data mining tool (Rittho, Klinkenberg, Fischer, Mierswa, & Felske, 2001). Each feature was tested against the 'best' label, and the resulting weights as presented in Figure 3, were used for the final selections. A set of 12 features for all of the three subsets of the main data were selected for the final study including: Curvature, Diff1_acf10, e_acf1, e_acf10, entropy, Seasonal_strength1, Seasonal_strength2, trend, X_acf1, X_acf10, Seas_acf1, and linearity. The main idea was to select the top features with the highest weights; however, one feature *'linearity'* was not amongst the highest cases for the top 4 forecaster dataset, and so it was deliberately added to the set of selected features. As a result, it meant a unique set of features would be used for all of the three main subsets of the dataset and the analysis would be more comprehensible. More details about the OneR method can be found in Barak and Modarres (2015). The goal of using the feature selection method was to find the most effective meta-features and to reduce the dimensionality of the inputs for the meta-learners in order to achieve better accuracies and forecasting errors.

## 4.5 New features from Principle component analysis (PCA)

One other approach for dimensionality reduction was the PCA. Although this method rotates the whole feature space and mixes the original data, it can be useful for studying the sole effects of the number of dimensions and the covariance in the feature space (Figure 2) on the meta-learners. Here three principal components with a cumulative proportion of 99.961% were selected.

## 4.6 Results for various subsets of forecasting algorithms

In this part, we analysed the results obtained by meta-learning algorithms on four, six and seven different subsets of forecasting algorithms. For this purpose, forecast errors of the meta-learning algorithms were compared with individual forecasting algorithms in different cases. Based on the features extracted in section 4.3, six different classifiers were trained on the training data, and then the meta-learning forecast errors were obtained for the test data. Here, a 20%-80% ratio was specified for the test/training data with a 5-times repeated 10-fold cross-validation. Additionally, the forecast errors were calculated after implementing feature selection methods and principal component analysis. In Tables 3, 4, and 5, comparisons of two forecast error measures for meta-learning algorithms and individual forecasting algorithms are presented. Here, the boundary value is the least possible value of the desired error measure for a given test data. Based on the random variable characteristics of the time series, it is not possible to obtain the boundary errors (we only indicate it here as an ultimate threshold). As a result, three different boundary values were specified for each of the main subsets of the preliminary data.

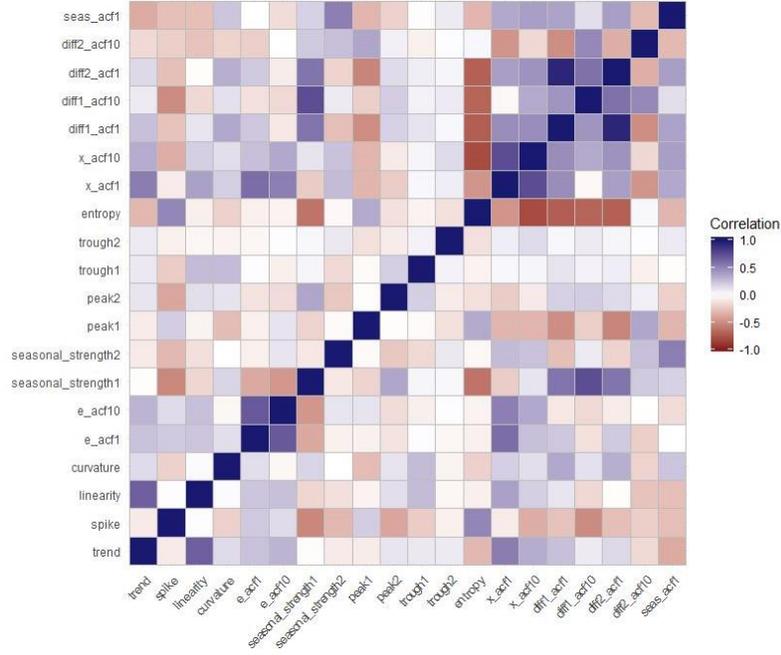

Figure 2. The covariance matrix for the 21 non-constant features

Table 3. A comparison of individual forecasting algorithms and meta-learners for top 4 forecasting models

| Model name | sMAPE | sMAPE (Feature selection) | sMAPE (PCA) | MASE | MASE (Feature Selection) | MASE (PCA) |
|---|---|---|---|---|---|---|
| **Boundary** | 0.204331 | 0.204331 | 0.204331 | 0.855394 | 0.855394 | 0.855394 |
| **Decision Tree** | 0.208788 | 0.209754 | 0.211926 | 0.890636 | 0.890848 | 0.898636 |
| **Neural Network** | 0.217160 | 0.210782 | 0.213192 | 0.896681 | 0.885106 | 0.896681 |
| **Random Forest** | 0.209063 | 0.208708 | 0.214312 | 0.875621 | **0.875181** | 0.904848 |
| **SVM** | 0.210506 | 0.208511 | 0.211925 | 0.877244 | 0.884090 | 0.899848 |
| **Treebag** | 0.211038 | **0.207519** | 0.215966 | **0.874272** | 0.891941 | 0.907424 |
| **XGboost** | **0.208324** | 0.210691 | 0.211926 | 0.876424 | 0.885848 | **0.891666** |
| MAPA | 0.210937 | 0.210937 | **0.210937** | 0.893030 | 0.893030 | 0.893030 |
| TBATS | 0.217160 | 0.217160 | 0.217160 | 0.903439 | 0.903439 | 0.903439 |
| Thief | 0.213300 | 0.213300 | 0.213300 | 0.896681 | 0.896681 | 0.896681 |
| ZZZ | 0.211573 | 0.211573 | 0.211573 | 0.892909 | 0.892909 | 0.892909 |

In Table 3, highlighted numbers in each column represent the lowest error obtained for the corresponding forecasting method. Before implementing feature selection and PCA, meta-learning algorithms have outperformed the individual forecasting methods for both measures of sMAPE and MASE.

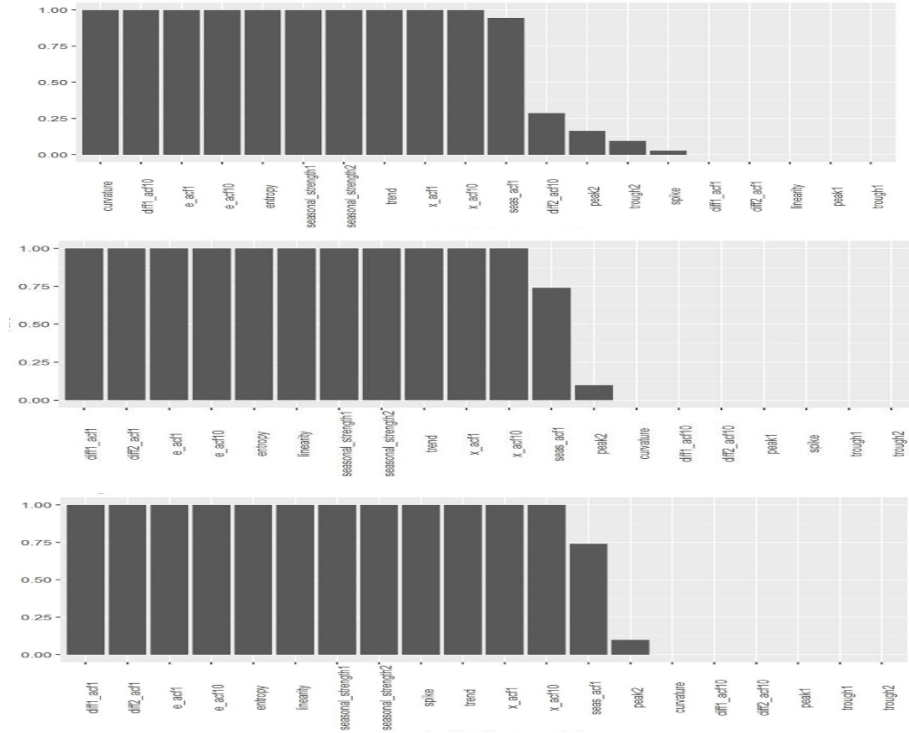

Figure 3. Weights of different features achieved by OneR for the 4(top), 6 (middle), and 7 (bottom) forecasters

The minimum sMAPE obtained by individual forecasting algorithms is 0.210937, which is higher than the value of sMAPE for the decision tree, random forest, SVM, and XGboost. For MASE, ZZZ was the best performer among the individual forecasters; however, this forecaster was outperformed by all the meta-learners.

Furthermore, the feature selection approach obtained a better forecasting performance in comparison to individual methods. However, after dimensionality reduction with PCA, the performance of meta-learning algorithms deteriorated in a way that for the sMAPE, the lowest error belongs to an individual forecasting algorithm. Figure 4 visualises the meta-learning algorithm forecast errors before and after implementing dimensionality reduction.

Regarding the sMAPE in Figure 4 (right-hand side), the meta-learning forecast errors for the neural network and support vector machine have been decreased by the implementation of the feature selection. Among the tree base methods, random forest and bagged tree have performed better after the feature selection. However, after the implementation of PCA, only the forecast error of neural networks has decreased, and the performances of other meta-models have deteriorated. In the MASE case (left-hand side of Figure 4), after implementing feature selection methods, the forecast errors of the neural network and random forest have decreased. After PCA, none of the meta-learning algorithms have been improved.

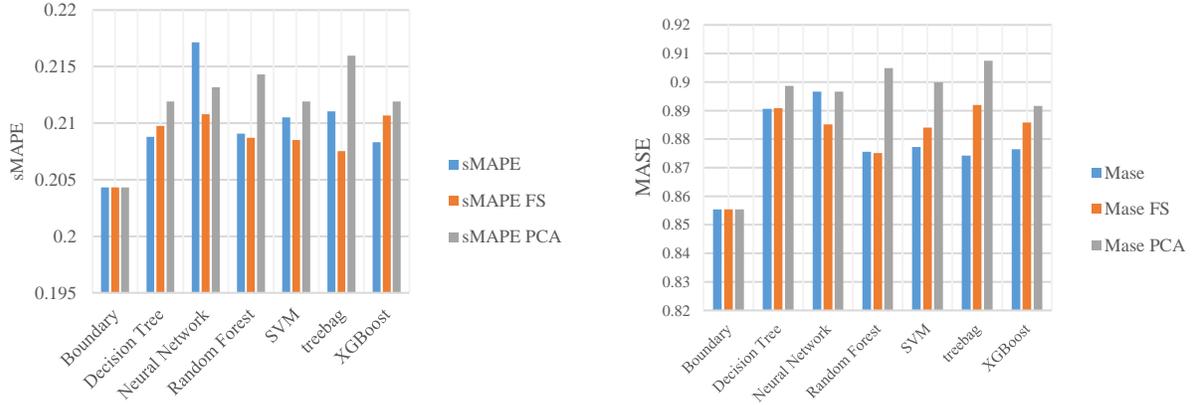

Figure 4. Performance of each meta-learning algorithms for the top 4 forecasters with sMAPE and MASE

In Table 4, the errors of meta-models with six top forecasting algorithms before and after implementing the dimensionality reduction methods, are presented. Again, in each column, highlighted numbers represent the lowest error made by an algorithm for the corresponding error measures. Before implementing the dimensionality reduction methods, for the sMAPE measure, four of the meta-learners outperformed the best individual forecaster (MAPA). Neural network and SVM were not able to surpass all the individual forecasters. For the MASE measure, four out of six meta-learners performed better than individual forecasters. Neural network and SVM were the only meta-learners with a higher MASE value than the best individual forecaster (ZZZ).

After implementing PCA, the performances deteriorated considerably. The result of the feature selection is the opposite, and the errors of meta-learners have decreased. For more illustration, the forecast errors are presented in the Figure 5.

Table 4. A comparison of individual forecasting algorithms and meta-learners for the top 6 forecasting models

| Model name | sMAPE | sMAPE (Feature selection) | sMAPE (PCA) | MASE | MASE (Feature Selection) | MASE (PCA) |
|---|---|---|---|---|---|---|
| **Boundary** | 0.203075 | 0.203075 | 0.203075 | 0.849636 | 0.849636 | 0.849636 |
| **Decision Tree** | **0.208198** | **0.208249** | 0.213146 | 0.879878 | 0.879121 | 0.898378 |
| **Neural network** | 0.217160 | 0.209909 | 0.213836 | 0.903439 | 0.888015 | 0.903439 |
| **Random Forest** | 0.209052 | 0.208970 | 0.214647 | **0.872136** | **0.873196** | 0.901833 |
| **TreeBag** | 0.208314 | 0.209152 | 0.215957 | 0.879272 | 0.888015 | 0.902636 |
| **SVM** | 0.213179 | 0.210401 | 0.211010 | 0.885545 | 0.888015 | **0.889681** |
| **XGboost** | 0.208318 | 0.208920 | **0.210937** | 0.903439 | 0.891727 | 0.903439 |
| DOTM | 0.215662 | 0.215662 | 0.215662 | 0.906030 | 0.906030 | 0.906030 |
| MAPA | 0.210937 | 0.210937 | **0.210937** | 0.893030 | 0.893030 | 0.893030 |
| STheta | 0.215075 | 0.215075 | 0.215075 | 0.903151 | 0.903151 | 0.903151 |
| TBATS | 0.217160 | 0.217160 | 0.217160 | 0.903439 | 0.903439 | 0.903439 |
| Thief | 0.213300 | 0.213300 | 0.213300 | 0.896681 | 0.896681 | 0.896681 |
| ZZZ | 0.211573 | 0.211573 | 0.211573 | 0.892909 | 0.892909 | 0.892909 |

The left-hand side of Figure 5 for the sMAPE measure can reveal that both the feature selection and PCA could lower the error of the neural network and support vector machine algorithms. By doing this comparison for tree-based methods, it can be understood that the feature selection has only decreased the errors of random forest, whilst PCA has led to an increase in error of all tree-based algorithms. From the right-hand side of the plot for the MASE measure, it seems the feature selection has led to a decrease in the errors of the neural network and decision tree; the other meta-learners have experienced an increase in their errors. Also, after PCA, none of the errors have been improved.

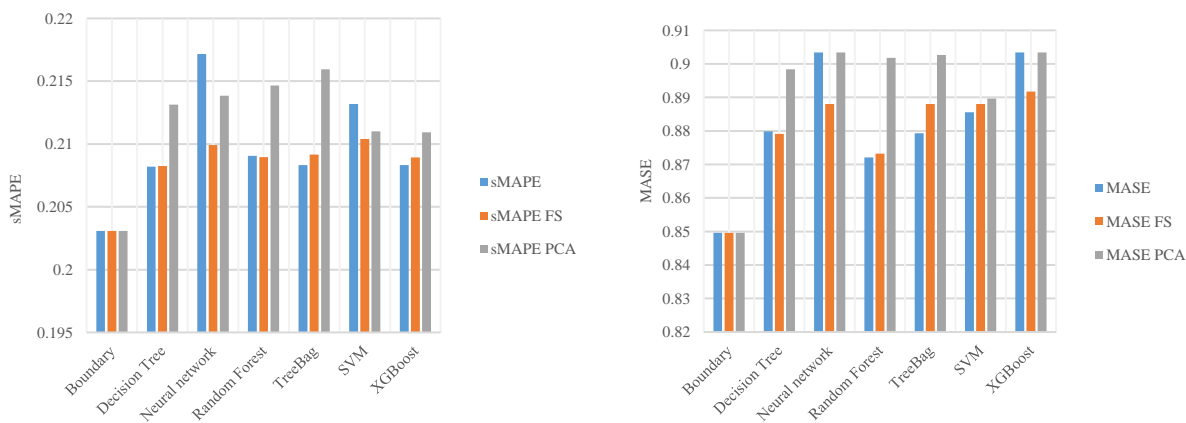

Figure 5. A comparison of forecast errors before and after implementation of feature selection and PCA

Finally, in Table 5, the same computations with seven top forecasting algorithms are analysed. Before implementing dimensionality reduction, three of the meta-learners outperformed all individual forecasters for sMAPE. For MASE, the neural network is the only meta-learners that was not able to surpass all of the individual forecasters.

Table 5. A comparison of individual forecasting algorithms and meta-learners for 7 forecasting models

| Model name | sMAPE | sMAPE (Feature selection) | sMAPE (PCA) | MASE | MASE (Feature Selection) | MASE (PCA) |
|---|---|---|---|---|---|---|
| Boundary | 0.201585 | 0.201585 | 0.201585 | 0.836788 | 0.836788 | 0.836788 |
| Decision Tree | 0.212638 | 0.213615 | 0.211779 | **0.874303** | 0.881727 | 0.893030 |
| Neural network | 0.217160 | 0.208898 | 0.210580 | 0.903439 | 0.881469 | **0.891010** |
| Random Forest | 0.210309 | 0.210364 | 0.217639 | 0.878909 | **0.877787** | 0.901318 |
| SVM | 0.212737 | 0.209316 | **0.209790** | 0.889272 | 0.888196 | 0.893030 |
| TreeBag | **0.208682** | 0.210223 | 0.213938 | 0.888575 | 0.884651 | 0.901545 |
| XGboost | 0.210530 | **0.208791** | 0.210937 | 0.883545 | 0.893030 | 0.893030 |
| DOTM | 0.215662 | 0.215662 | 0.215662 | 0.906030 | 0.906030 | 0.906030 |
| MAPA | 0.210937 | 0.210937 | 0.210937 | 0.893030 | 0.893030 | 0.893030 |
| Snaive | 0.252219 | 0.252219 | 0.252219 | 1 | 1 | 1 |
| STheta | 0.215075 | 0.215075 | 0.215075 | 0.903151 | 0.903151 | 0.903151 |
| TBATS | 0.217160 | 0.217160 | 0.217160 | 0.903439 | 0.903439 | 0.903439 |
| Thief | 0.213300 | 0.213300 | 0.213300 | 0.896681 | 0.896681 | 0.896681 |
| ZZZ | 0.211573 | 0.211573 | 0.211573 | 0.892909 | 0.892909 | 0.892909 |

Figure 6 represents a comparison of the performance of six meta-learning algorithms, with seven forecasting algorithms for the sMAPE and MASE as error measures, before and after implementing the feature selection and PCA. Similar to the meta-learning with four and six forecasting algorithms, the feature selection has led to a considerable decrease in the sMAPE of the neural network. For tree-based algorithms, the forecast errors have increased after feature selection. For the MASE measure, the neural network's improvement is apparent. For the random forest, SVM, and bagged tree the feature selection improves the forecast error, and for the rest, the results have worsened. PCA has not led to a decrease in the forecast error in any of the models.

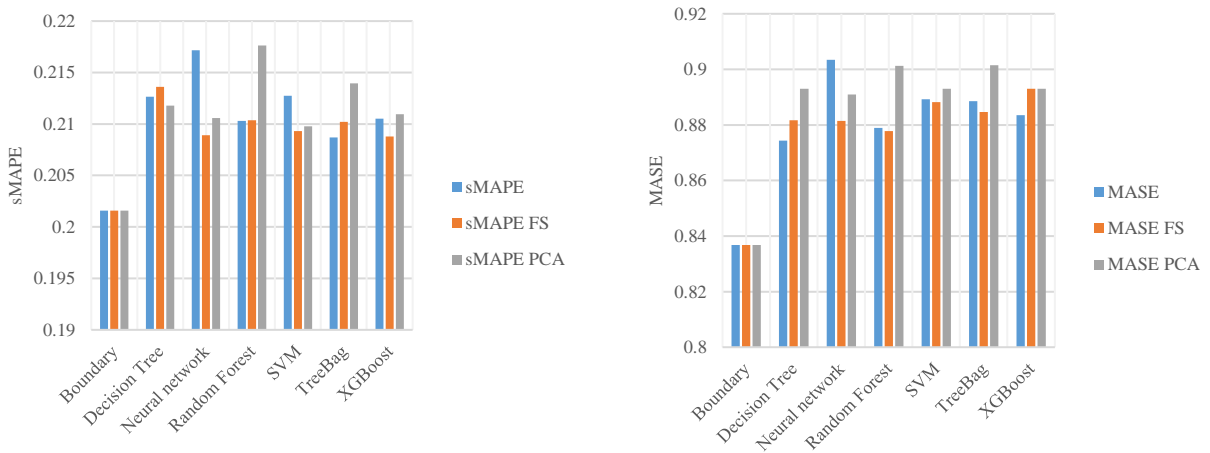

Figure 6. The forecast errors (sMAPE and MASE) of meta-learning with seven forecasting algorithms before and after implementing feature selection and PCA

## 4.7 Analyzing the effect of the number of forecasters

After implementing different meta-learning algorithms, the results between the three main sub-branches of the 'preliminary data' (top 4, top 6, and top 6 with Snaive) could be compared. For each subset of forecasters, most of the meta-learners outperformed the top results obtained by individual forecasters for both sMAPE and MASE. Among the meta-learners, the neural network had the weakest performance and was not able to outperform the best individual forecasters for each subset.

In Figure 7, a comparison of errors (sMAPE and MASE) achieved by the meta-learners for different subsets of forecasting algorithms, is presented. It can be seen in the left-hand side of Figure 7, sMAPE measures that go from 4 to 6 forecasting algorithms, have led to a decrease in the forecast errors of the decision tree, SVM, XGboost, and random forest; however, adding Snaive to the list of forecasters, has led to an increase in the errors of meta-learners. For the neural network, forecast errors are independent of the number of forecasting algorithms, and for the bagged decision tree (Treebag), a certain pattern cannot be recognized from varying the subset of the forecasting model. With the MASE as the error measure, only the performance of decision tree and random forest have improved from four to six individual forecasting algorithms. From six to seven forecasters, adding seasonal naïve has led to an increase in the meta-learning errors except for decision tree and XGboost.

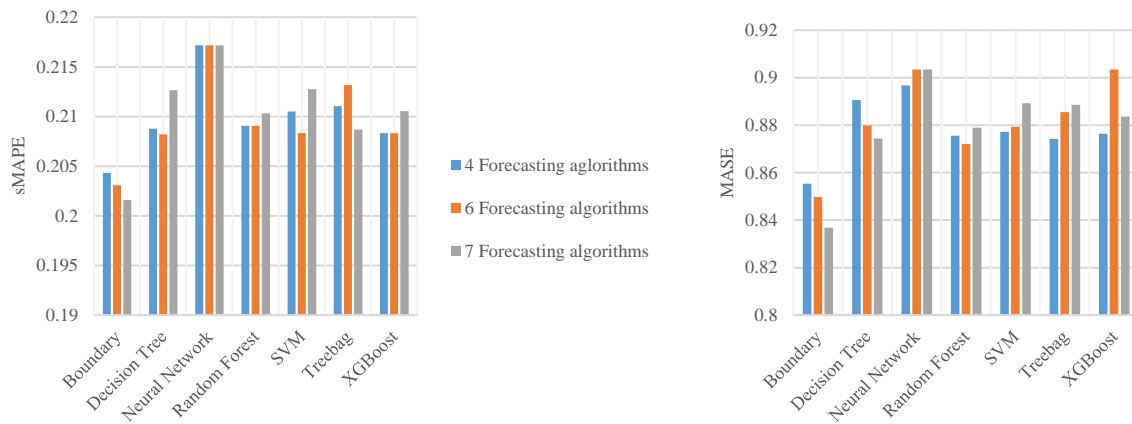

Figure 7. The forecast errors (sMAPE and MASE) of meta-learning with different subsets of forecasters

## 4.8 Meta-learning on basic forecasters

In this part, we tested the results of applying meta-learning-based recommender system on a pool of basic models. Table 6 represents a comparison of forecast errors for basic forecasters and meta-learners. It can be noticed that in general, most of the meta-learners outperformed the basic models for both sMAPE and MASE. For sMAPE, all of the meta-learners except for Treebag outperformed the individual forecasters. For the case of MASE, XGboost and random forest outperformed all the individual forecasters. Treebag, decision tree, and neural network defeated three of the four individual forecasters and SVM could only surpass two of the four basic forecasters.

This experiment has proven that even in basic forecasting models, which are mainly used in business for their simplicity and speed of computation time with large data, the meta-learning approach can enhance the accuracy of the results and enhance their efficiency in the general forecasting community.

It is worthy to note that using the NN5 data set in the paper showcases the efficiency of the idea while the process for the further dataset can be replicable with the same procedure.

Table 6. A comparison of basic forecasters and meta-learners based on sMAPE and MASE

| Model name | sMAPE | MASE |
|---|---|---|
| Boundary | 0.209369 | 0.878758 |
| Decision Tree | 0.211402 | 0.892909 |
| Neural Network | 0.211573 | 0.892909 |
| Random Forest | **0.210904** | 0.891757 |
| SVM | 0.211729 | 0.894303 |
| Treebag | 0.211915 | 0.892893 |
| XGboost | 0.211729 | **0.890878** |
| ANN | 0.365938 | 1.680030 |
| AAN | 0.367728 | 1.688878 |
| ANA | 0.211729 | 0.893333 |
| AAA | 0.212965 | 0.892181 |

## 5. Discussion

In this paper, six different classifiers were utilised to select the best forecasting algorithms for the different time series of the NN5 competition dataset. At first, thirteen forecasting methods were implemented; however, two of the forecasters (ARIMA and bagged tree) exhibited non-consistent behaviours. Hence, we decided to discard these two forecasters. One other important fact about our forecasters is the presence of the ZZZ method. This method uses the AIC measure to choose the best model amongst a set of exponential smoothing methods. As a result, we have indirectly applied an AIC-based model selection approach in our recommender system schemes.

The performances of each of the meta-learners were examined through changing the number of forecasting methods and error measures. Three subsets consisting of the top 4, top 6, and top 6 with the seasonal naïve method (7 models) were used as a pool of input for meta-learning. The true distribution of the 'best' labels for the test data has been provided in Table 7. To escape from a possible effect of the curse of dimensionality in this paper, we applied two main dimensionality reduction methods, namely PCA and feature selection.

To have a better understanding about the underlying processes of the meta-learners, we analysed the outputs of meta-learners by focusing on their resulting confusion matrices. For the case of meta-learners without dimensionality reduction, all of the meta-learners, except for the decision tree and Treebag, have never recommended the ZZZ method for any of the time series in the test data for 4,6 and 6 with Snaive cases. Furthermore, the decision tree and Treebag only suggested the ZZZ model in the top 4 and 6 datasets, and there was no ZZZ label for the 6 with Snaive data. Also, for all of the meta-learners except for the neural network, 80% of the recommended outputs were labelled as the MAPA or the Thief methods; hence the meta-learners would tend to choose the most powerful methods for the majority of the test data. This tendency toward the MAPA and Thief methods was observed for all of the top 4, 6 and 6 with Snaive cases. However, in the case of the neural network, this meta-learner recommended the TBATS method for most of the test data.

After feature selecting by the *OneR* algorithm, all of the meta-learners tended to diversify their choices for the output labels. The extreme case for this diversification was observed for the neural networks, where a drastic improvement could be seen in their performances as a result of this change. This diversification had a downside for some of the meta-learners. In some cases, where the classification accuracy of the meta-learners for the before and after feature selection had not changed, the suggested labels shifted from being majorly the most powerful forecasters such as MAPA and Thief, to weaker ones.

PCA deteriorated the performances of almost all of the meta-learners. After further investigation, it was revealed that the data had transformed in such a way that meta-learners tended to exhibit the same diversification in their output labels as discussed above; however, this time with less classification accuracy. It seems the quality of the generated data under PCA was much weaker than the feature selection. Feature selection methods led to better results, whereas implementing PCA had a negative impact on the performance of classifiers. This negative impact was most noticeable for the tree-based methods. However, PCA enhanced the performance of the neural network classifier for both sMAPE and MASE (see Tables 3, 4 and 5).

Table 7. Distribution of the labels in the test data – each number represents a percentage value

|        | sMAPE | | | MASE | | |
|---|---|---|---|---|---|---|
|        | 4 | 6 | 7 | 4 | 6 | 7 |
| MAPA   | 18.18 | 9.09  | 13.64 | 13.64 | 13.64 | 13.64 |
| THIEF  | 22.73 | 13.64 | 9.09  | 27.27 | 13.64 | 9.09  |
| ZZZ    | 31.82 | 27.27 | 27.27 | 27.27 | 22.73 | 18.18 |
| TBATS  | 27.27 | 27.27 | 22.73 | 31.82 | 31.82 | 22.73 |
| STheta | -     | 13.64 | 13.64 | -     | 4.55  | 9.09  |
| DOTM   | -     | 9.09  | 9.09  | -     | 13.64 | 9.09  |
| Snaive | -     | -     | 4.55  | -     | -     | 18.18 |

Tables 8 and 9 represent the percentage of individual forecasting algorithms outperformed by each of the meta-learners for both sMAPE and MASE. It seems that before implementing feature selection and PCA, most of the meta-learners produced better results than the individual forecasters. After applying feature selection, almost all of the meta-learners surpassed the whole individual forecasting algorithms in different subsets of forecasters. For the top 4 and top 6 forecast methods input, feature selection led to a better outperformance percentage for all of the meta-learners in both sMAPE and MASE measures. For 7 forecasters input with sMAPE as the error measure, the feature selection led to a decrease in outperformance percentage of the decision tree, and for the case of MASE, the feature selection decreased the outperformance percentage of XGboost.

According to Tables 8 and 9, the implementation of the PCA deteriorated the performances of almost all of the meta-learners. However, the neural network meta-learners experienced a drastic increase in their performances with the PCA, and they even outperformed all of the individual forecasters for both sMAPE and MASE. The top scores in each category of Tables 8 and 9 are shown in bold numbers.

Table 8. The percentage of individual forecasters outperformed by each of the meta-learners for sMAPE

| Meta learners | 4 | 6 | 7 | 4 FS | 6 FS | 7 FS | 4 PCA | 6 PCA | 7 PCA |
|---|---|---|---|---|---|---|---|---|---|
| Decision Tree | **100** | **100** | 71 | **100** | **100** | 57 | **50** | 67 | 71 |
| Neural Network | 0 | 0 | 14 | **100** | **100** | **100** | **50** | 50 | **100** |
| Random Forest | **100** | **100** | **100** | **100** | **100** | **100** | 25 | 50 | 14 |
| SVM | **100** | 67 | 71 | **100** | **100** | **100** | **50** | **83** | **100** |
| Treebag | 75 | **100** | **100** | **100** | **100** | **100** | 25 | 17 | 57 |
| XGboost | **100** | **100** | **100** | **100** | **100** | **100** | **50** | **83** | 85 |

Table 9. The percentage of individual forecasters outperformed by each of the meta-learners for MASE

| Meta learners | 4 | 6 | 7 | 4 FS | 6 FS | 7 FS | 4 PCA | 6 PCA | 7 PCA |
|---|---|---|---|---|---|---|---|---|---|
| Decision Tree | **100** | **100** | **100** | **100** | **100** | **100** | 25 | 50 | 71 |
| Neural Network | 25 | 17 | 28 | **100** | **100** | **100** | 25 | 17 | **100** |
| Random Forest | **100** | **100** | **100** | **100** | **100** | **100** | 0 | 50 | 57 |
| SVM | **100** | **100** | **100** | **100** | **100** | **100** | 25 | **100** | 71 |
| Treebag | **100** | **100** | **100** | **100** | **100** | **100** | 0 | 50 | 57 |
| XGboost | **100** | 17 | **100** | **100** | **100** | 71 | **100** | 17 | 71 |

Another important issue was the decreasing errors in the boundary values from the 4 to 7 pool of forecasters. Table 10 presents the number of classifiers that were able to detect the decreasing pattern in sMAPE and MASE. Before implementing the feature selection method and PCA for sMAPE, none of the meta-learners were able to detect the aforementioned decreasing pattern, and for MASE, the pattern was detected only by the decision tree algorithm. After feature selection, two of the meta-learners (neural network and XGboost) detected the pattern for sMAPE, and one meta-learner (Treebag) detected the pattern for MASE. After PCA the recognition of the pattern was not done for any of the meta-learners for sMAPE; however, two of the meta-learners (decision tree and random forest) were able to find the pattern for MASE.

Table 10. Number of meta-learners detected the decreasing pattern

| Different cases | sMAPE | MASE |
|---|---|---|
| Pure meta learning | 0 | 1 |
| Meta learning with Feature selection | 2 | 1 |
| Meta learning with PCA | 0 | 2 |

Although most of the meta-learners were able to detect the decreasing value of sMAPE and MASE from 4 to 6 forecasting algorithms, this decrease in error measures was not recognised well from 6 to 7 forecasters. One reason for this could be the presence of the Snaive method amongst the seven forecasting algorithms. It seems the addition of the Snaive method with its weak performance, meant the meta-learners found it difficult to discover the pattern. Another possible reason could be the lack of sufficient data since the NN5 dataset contained only 111-time series.

For the case of the neural networks, it is expected that with the use of a multilayer neural network, the weights assigned by the network itself will take care of the dimensionality as in Hinton and Salakhutdinov (2006); however, since the networks used here were multilayer perceptron (MLP) with three layers, there was a suspicion regarding this ability of our networks. As depicted in Tables 8 and 9, the neural network has benefitted the most from dimensionality reduction.

Since all our tree-based methods (bagged Tree, Decision tree, XGboost, and random forest) have intrinsic built-in feature selection methods to assign variable importance to the input attributes, these algorithms tend not to use all of the feature space for their classification. This behaviour was one of the main motivators for applying the dimensionality reduction methods to study the general effects on all of the meta-learners. There have been numerous works lately on how to address the built-in feature selection of the tree-based method, for more details readers can refer to Nguyen, Huang, and Nguyen (2015).

One important discovery after applying the dimensionality reduction was the change in performance of the meta-learners with regards to different error measures. It seems the performances of the learners with bootstrap aggregation in their algorithms, deteriorated after applying dimensionality reduction for the sMAPE error measure. However, this pattern was not detected in the MASE. Therefore, it implicitly demonstrates the robustness of the MASE for the meta-learning studies.

## 6. Conclusion

In this study, we investigated three new approaches in using meta learning for model selection in time series forecasting. For the first time, we used different subsets of forecasting methods as the outcome attributes for meta-learning to analyse the effect of a number of forecasters in the input pool. For six different meta-leaners, we analysed the behaviour of each meta-learner and considered different subsets of the individual forecaster in the pool of forecasters. In normal situations, where all of the individual forecasters used in meta-learning are chosen from a homogenous set of powerful methods, it was observed that adding more individual forecasters to the pool improved the performance of meta-learners. We also tried to do a sensitivity analysis by adding a weak forecaster to the pool of forecasters. For this reason, we evaluated the behaviour of meta-learners by adding Snaive as a weak forecaster to the input pool. One lesson learned from this was that although adding a weak forecaster may lead to a decrease in boundary value, it will cause a deterioration in the performances of meta-learners.

Another aim of this work was taking advantage of two different error measures: sMAPE and MASE. One result obtained was that MASE is more suitable for the decision tree, RF, and Treebag, whilst the sMAPE was better for the neural network and XGboost in our scheme.

We investigated the role of dimensionality reduction by implementing feature selection methods and PCA. In general, feature selection had improved the errors of meta-learners and would help them outperform a higher percentage of individual forecasters. However, in our case, principal component analysis worsened the performances of meta-learners drastically, and as a result, it seems for this scheme, it is better to use feature selection instead of PCA for dimensionality reduction.

Finally, we analysed the performances of our model on a pool of forecasters related to business applications, known in this paper as basic forecasters. It was revealed that like our main pool of state-of-the-art forecasters, the meta-learners were able to crush the basic forecasters. These results indicate that the recommender systems model described in this paper for time series forecasting, can drastically improve and benefit not only advanced forecasting models, but also industry-level methods.

For future studies, implementation of the procedure on different datasets and evaluating the effect of the number of time series on the meta-learning results could be considered. Moreover, the accuracy of forecasting errors could be increased by using more developed ensemble meta-learners. Finally, combination of forecasting methods with meta-learning instead of model selection, might be promising.

**Acknowledgement**

The research was supported by the Czech Science Foundation (GACR Project 18-15530S) and Operational Program Education for Competitiveness – Project No. CZ.1.07/2.3.00/20.0296.